\documentclass[letterpaper]{article}
\usepackage{aaai}
\usepackage{times}
\usepackage{helvet}
\usepackage{courier}
\usepackage{xcolor}
\usepackage{adjustbox}
\usepackage{cleveref}
\usepackage{parskip}
\usepackage{tabularx, booktabs}
\usepackage[toc,page]{appendix}

\usepackage{natbib}
\usepackage{url}
\usepackage{algorithm}
\usepackage{algorithmic}

\usepackage{newfloat}
\usepackage{listings}
\floatstyle{ruled}
\newfloat{listing}{tb}{lst}{}
\floatname{listing}{Listing}

\begin{document}
%
\title{Hierarchical Multi-Label Classification of Online Vaccine Concerns}
\author{
    Chloe Qinyu Zhu\thanks{\textbf{Equal Contribution}}\\
    Duke University\\
    qinyu.zhu@duke.edu\\
    \And
    Rickard Stureborg$^*$\\
    Duke University\\
    rickard.stureborg@duke.edu\\
    \And
    Bhuwan Dhingra\\
    Duke University\\
    bdhingra@cs.duke.edu\\
}

\nocopyright 
\maketitle

\begin{abstract}
\begin{quote}
    Vaccine concerns are an ever-evolving target, and can shift quickly as seen during the COVID-19 pandemic.
    Identifying longitudinal trends in vaccine concerns and misinformation might inform the healthcare space by helping public health efforts strategically allocate resources or information campaigns.
    We explore the task of detecting vaccine concerns in online discourse using large language models (LLMs) in a zero-shot setting without the need for expensive training datasets.
    Since real-time monitoring of online sources requires large-scale inference, we explore cost-accuracy trade-offs of different prompting strategies and offer concrete takeaways that may inform choices in system designs for current applications.
    An analysis of different prompting strategies reveals that classifying the concerns over multiple passes through the LLM, each consisting a boolean question whether the text mentions a vaccine concern or not, works the best.
    Our results indicate that GPT-4 can strongly outperform crowdworker accuracy when compared to ground truth annotations provided by experts on the recently introduced VaxConcerns dataset, achieving an overall F1 score of $78.7\%$.

\end{quote}
\end{abstract}

\section{Introduction} 


Vaccine hesitancy is a critical topic in public health and identifying concerns contributing to hesitancy can provide opportunities for improving community health outcomes and resource allocation strategies.
Identifying societal trends in vaccine concerns can also inform healthcare interventions, such as responding to concerns by debunking or ``prebunking'' vaccine myths.

Much public health work has been focused on describing the landscape of misinformation and concerns surrounding vaccination.
One such work introduces VaxConcerns, a disease-agnostic taxonomy of concerns that may drive people towards hesitancy \citep{stureborg2023interface}.
The VaxConcerns taxonomy organizes concerns into two levels, one of broad granularity with concern categories such as ``Health Risks'' and another of finer granularity with specific claim categories such as vaccines having ``Harmful Ingredients'' or ``Specific Side-Effects''.
VaxConcerns is composed of 5 parent categories and 19 child classes each associated with one (and only one) parent category.

One of the proposed tasks for VaxConcerns is to classify text (e.g. blog articles, tweets) into the taxonomy.
Tasks where multiple labels can be selected at once are referred to as multi-label classification.
Since the taxonomy is hierarchical, and multiple concerns can be brought up by a single passage of text, this constitutes a hierarchical multi-label classification task.
Hierarchical multi-label classification requires an independent binary (``present'' or ``not present'' in the text) prediction for every label in the taxonomy.
These predictions must be made separately for parent and child categories, since a text can invoke the broad category (e.g. Lack of Benefits) without a specific rationale (e.g. Existing Alternatives).
For example, consider the YouTube comment: ``\texttt{I don't need the vaccine! No reason to get it}'', which clearly invokes the parent Lack of Benefits without invoking any of the child labels.

Hierarchical multi-label classification requires careful design considerations when constructing classifiers, especially when using generative language models.
A \textit{fully labeled} example is a passage which has a binary (``present'' or ``not present'') prediction for every label in the taxonomy.
These predictions can be either batched together all at once (single-pass) or split into multiple passes (multi-pass) to the model.
Further, the choice of which groups to batch together introduces more options.
Labels can be shown on their own as a binary choice (binary), as a short flat list of options (multi) or as a tree-structure (hierarchical, or hrchl).
For the purpose of notation, we will combine the notion of passes and label combinations by combining these short-hand names using the convention defined in \citet{cartwright2019sound}, as further explained in the methodology.
Therefore, `multi-pass binary' denotes the case where every label is passed as a binary question to the model in parallel requests.

To build automatic classification systems for detecting vaccine concerns, we experiment on the dataset of anti-vaccination passages introduced by \citet{stureborg2023interface}.
The passages are labeled (mapped) to the VaxConcerns taxonomy by expert annotators in multiple rounds, reaching agreement among all experts through written dialogue.
This dataset is considered very high-quality, and is used by \citet{stureborg2023interface} to benchmark the quality of crowdsource worker annotations.
An example passage from the dataset is shown below:
    \begin{quote}
        ``\texttt{The Minister of Fear (the CDC) was working overtime
    peddling doom and gloom, knowing that frightened
    people do not make rational decisions - nothing sells
    vaccines like panic.}''
    \end{quote}

Using this dataset, we build automatic classifiers by prompting Large Language Models (LLMs), and investigate considerations such as (a) cost, (b) output consistency, (c) performance, and (d) prompting strategies.
We ultimately build a system that strongly outperforms the best crowdsource annotation by using GPT-4-Turbo, and find four Pareto-optimal system designs to optimize cost and performance.
We offer guidelines that may help public health efforts during design considerations of their automatic classifiers.

This work offers the following novel contributions:
\begin{itemize}
    \item Comparison of prompting strategies' impact on hierarchical multi-label classification performance
    \item Formalize \textit{Format Demonstrations} as an instruction in LLM prompting and measure its effectiveness in reducing output format errors
    \item Benchmark of various LLMs on this task, finding Llama-7B and Llama-13B models are not feasible due to format errors and GPT-4 with multi-pass binary prompting achieves SOTA performance of 78.75\% F1 score at temperature=0.
    \item OpenAI's GPT-4-Turbo model does not outperform its predecessor GPT-4. GPT-4 achieves 0.92\% higher absolute F1 score on average as compared to GPT-4-Turbo.
    \item Findings show our models have substantially better performance than the most accurate human-based crowdsource annotation baselines
    \item An analysis of cost limitations of models and prompting strategies for large scale classification
\end{itemize}

\section{Related work}

Hierarchical multi-label classification is a task where many labels can be selected at the same time, and the labels have a hierarchical structure (giving the task various granularities of labels).
It is a format used across many disciplines \citep{cartwright2019sound, zhang_ontological_2018, sadat2022hierarchical}.

Vaccine misinformation is well studied in public health, especially through a social media lens \citep{kaufman_comprehensive_2017, jacobson_taxonomy_2007, nuzhath_covid19_2020, cotfas_longest_2021, stureborg2023interface}.
Some of this work has pointed out that misinformation claims and concerns can evolve over time \citep{gerber_vaccines_2009}.
This poses a challenge for researchers putting together resources such as training datasets for vaccine misinformation, since the landscape of misinformation is ever-changing.

One potential solution are systems that can function in zero-shot settings, requiring limited resources to be created before being able to classify a new piece of misinformation.
Recent advances in Large Language Models (LLMs) have allowed high-accuracy classification for a diverse set of tasks in both zero-shot and few-shot settings \citep{guo_how_2023, he_annollm_2023}.
LLMs allow the resources necessary for classification of a new vaccine myth to be reduced to just a 1-2 sentence description of the myth, rather than the collection of a dataset of examples.

Previous work has focused on reducing the costs of human annotation for classification of vaccine misinformation, \citep{stureborg2023interface, muller_crowdbreaks_2019} and other work explores classification under a setting with large training datasets \citep{weinzierl2022vaccinelies, zarei2023vaxculture}, but there remains a gap regarding exploring LLMs as few- or zero-shot classifiers of vaccine misinformation, particularly with respect to the scalability and cost of such systems.

Some work explores LLM classifier costs, but often it is compared with the cost of human annotation \citep{gilardi_chatgpt_2023}.
There remains open questions regarding the tradeoff between prompting strategies and costs which become relevant when classifying the volumes of text encountered on mainstream social media platforms.

\section{Methods}

\subsection{Dataset of Anti-Vaccination Text}
We utilize the VaxConcerns dataset presented by \citet{stureborg2023interface}.
This dataset contains 4,800 passage-label pairs across 200 randomly sampled passages from articles from anti-vaccination blogs online.
The anti-vaccination blogs do not always contain passages regarding vaccines, but more than half of the passages are labeled with at least one concern in the ground-truth labels.
Passages are approximately 2-5 sentences long, and were originally determined using HTML divider elements indicating paragraph breaks.

This dataset is labeled by experts in three rounds.
First, experts give map each passage to the relevant labels in the VaxConcerns taxonomy \citet{stureborg2023interface}, and any immediate agreements are set aside as `finished' labels.
Importantly, it is not a set of passages but rather passage-label pairs.
This is because the experts may agree on one of the labels (such as Harmful Ingredients) for a single passage, but disagree about a different label (such as Profit Motives).
We later refer to this group of labels as the \textit{Easy} partition, since experts did not have to go through any further steps to reach agreement on the ground-truth label.
Experts then went through a second round of annotation on the examples with disagreement, where they provided written rationales for why the passage should be labeled as either True or False for a given label in the VaxConcerns taxonomy.
Next, experts discussed these rationales and came to a consensus or voted by majority.
For the purpose of our experiments, we only separate the first round into the \textit{Easy} partition, and all other passage-label pairs are part of the \textit{Non-Easy} partition, since otherwise the size of these sets are too small to make significant claims.
We use the provided ground-truth labels as is, and do not perform any relabeling.

The previous work done by \citet{stureborg2023interface} also includes experiments regarding prompting strategies for crowdsource workers.
They conducted these experiments by asking crowdsource workers to annotate their dataset and measured performance under various interface designs by reporting the crowdsource workers' F1 scores.
We compare our experimental results with LLMs to these scores where relevant.

\subsection{Classification Models}
To build automatic classification systems, we use Large Language Models (LLMs).
LLMs have recently shown a high degree of accuracy across many tasks, even under zero-shot settings.
We investigate five models in this work:
\begin{itemize}
    \item GPT-3.5-Turbo, checkpoint \texttt{gpt-3.5-turbo-0613}
    \item GPT-4, checkpoint \texttt{gpt-4-0613}
    \item GPT-4-Turbo, checkpoint \texttt{gpt-4-1106-preview}
    \item Llama-2-7b, checkpoint \texttt{llama-2-7b-chat-hf}
    \item Llama-2-13b, checkpoint \texttt{llama-2-13b-chat-hf}
\end{itemize}
We investigated larger models (such as Llama-2-70b) but did not have enough compute at the time of the experiments to run this size model.

To prompt the models to give us accurate predictions, we use the following system and user prompts:
\begin{quote}
    \texttt{You are a healthcare expert. Your job is to find out vaccine related concerns from a given PARAGRAPH. You will be given a map of all concerns, where the key is as VaxConcerns\_1.1 and values are the actual concerns. Here’s the CONCERN MAP: \{concerns\} Please read the PARAGRAPH and tell me whether the concern is mentioned in the PARAGRAPH one by one. In your response, only return a map, same length as the CONCERN MAP, with the key exactly the one in the map, values as Yes or No.}
\end{quote}
We replace \texttt{\{concerns\}} with the relevant target labels according to one of the prompting strategies described in the next section.

We formalize the notion of Format Demonstrations, which are example outputs in the exact format we expect the model to produce.
These demonstrations are included in order to reduce errors in output formatting, as discussed further in Results.
In our case this demonstrates every key-label pair in the dictionary we want the LLM to return for classification.
For example, the below format demonstration is added to multi-pass multi during the first stage of annotation (when predicting level 1 labels only):
\begin{quote}
    \texttt{Here is an example output: \{"VaxConcerns 1": "Yes", "VaxConcerns 2": "No", "VaxConcerns 3": "No", "VaxConcerns 4": "Yes", "VaxConcerns 5": "No"\}}
\end{quote}
This gives the model explicit tokens to attend over to guide the format of the output generation, rather than just instructing the model we want a dictionary output.

Finally, the user prompt simply reads: \texttt{PASSAGE: \{passage\}} and the request is sent to the LLM, allowing it to generate a prediction as its response.
Predictions are always done in this zero-shot setting.
No predictions are made such that the context includes a previous prediction by the model (i.e. the ``chat history'' is cleared for each request).

\subsection{Prompting Strategies}
\begin{figure*}[t]
    \centering
    \includegraphics[width=\linewidth]{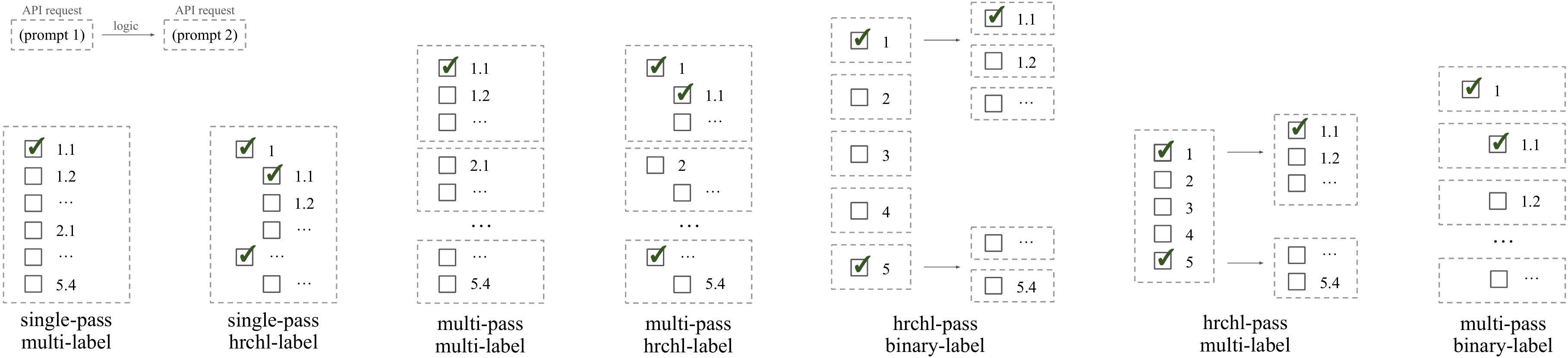}
    \caption{\textbf{Prompting Strategies Workflows} We experimented with 7 combinations of prompting and logic-passing style. Each prompting strategy is outlined above. Boxes represent stand-alone API calls to the language model, such as a single OpenAI request. The content of the box describes what final labels are requested in the respective prompt. Arrows show any logic that is carried out across these API requests, which only occurs if using a hierarchical-pass strategy.}
    \label{fig:prompting_workflow}
\end{figure*}

In order to produce a full prediction for a passage of interest (that is, a binary decision for every label in the taxonomy) we consider 7 potential prompting strategies.
This section describes each prompting strategy and introduces a short-hand notation used in the rest of this paper.
For each prompting strategy, we provide a visualization of how multiple passes are combined to get a fully labeled prediction (see \Cref{fig:prompting_workflow}).

\begin{itemize}
    \item Single-pass request of hierarchical-structured labels (\textbf{single-pass hrchl})
    --- This prompting strategy sends one request per passage to be labeled. In that request, the whole taxonomy is presented, and the model is asked to assign predictions for every label in the taxonomy, both within Level 1 and Level 2.
    \item Single-pass request of multiple labels (\textbf{single-pass multi})
    --- This prompting strategy sends one request per passage to be labeled, and asks the model to assign predictions to each of the Level 2 labels in the taxonomy. Unlike single-pass hrchl, it does not ask for explicit predictions on Level 1 labels. Instead, level 1 labels are inferred by whether or not a child label is present. This means Level 1 labels can only be present if the children are predicted present, which is an incorrect assumption of the ground-truth labels and therefore disadvantages this prompting strategy.
    \item Multiple parallel passes of requests with hierarchical-structured labels (\textbf{multi-pass hrchl})
    --- This strategy batches the prediction task into a few parallel requests (one for each parent and its children). It splits the taxonomy along parent categories and presents smaller trees to the model for labeling. For example, the first request would include the first Level 1 label, along with all of that label's children.
    \item Multiple parallel passes of requests with multiple labels (\textbf{multi-pass multi})
    --- Unlike multi-pass hrchl, this strategy omits the level 1 labels from explicit prediction. Instead, the first request includes only the children of the first Level 1 concern in the taxonomy. From these labels, the parent is inferred by whether or not any of the child labels are selected, similar to the logic applied in single-pass multi.
    \item Multiple parallel passes of requests with binary labels (\textbf{multi-pass binary})
    --- This strategy batches the annotation up even further by sending a unique request for every single label in the taxonomy. This achieves true independence among all of the labels, but requires far more requests and is therefore more computationally expensive.
    \item Multiple hierarchical passes of requests with multiple labels (\textbf{hrchl-pass multi})
    --- Hierarchical passes attempt to reduce the amount of necessary requests while still batching the prediction task into smaller chunks. It accomplishes this by first requesting predictions for all of the parent (Level 1) labels in a single request. Then, based on the predictions of this round, the strategy makes further requests for predictions of the children only if the parent has been indicated to be present.
    \item Multiple hierarchical passes of requests with binary labels (\textbf{hrchl-pass binary})
    --- This strategy performs the same hierarchical passes described above, but does so on each label individually. Therefore, there will always be at least five requests (one for each of the Level 1 concerns). If a request comes back with a positive prediction, this will launch N new requests, where N is the number of children for the parent predicted as positive.
\end{itemize}
\Cref{fig:prompting_workflow} shows useful workflow diagrams of these prompting strategies.

\subsection{Handling Generated LLM Predictions}
Once the model produces its predictions according to the prompting described in the above sections, we combine each request into a single dictionary of predictions, with one value for each concern in the VaxConcerns taxonomy.
When outputs are incorrectly formatted (i.e. they do not resolve to valid dictionaries in Python), or there are missing concerns from the total predictions made, we record these cases and retry the request.
We retry a maximum of 10 times before declaring the prediction as a failure to limit experimental costs.
Importantly, we only retry when the temperature parameter is set to a non-zero value, since otherwise we will sample a deterministic output and always return the same invalid prediction.

We record the cost of annotating each passage by counting up the number of input and output tokens by the models, and multiplying these by the cost of the model used. 
An average is taken over the total number of passages (200) to get the Total Cost per Passage.

We then determine F1 scores against the expert-determined ground-truth values using a macro-average, and record this value.

\section{Results}

We benchmark models on several dimensions to 
    determine the best prompting strategy 
    and cost considerations useful in order to scale the classification to social media-level quantities, and 
    determine the state-of-the-art (SOTA) performance on the VaxConcerns dataset.

\subsection{Failure Rate Analysis}

Language models, despite their advanced capabilities, are susceptible to a range of errors that can manifest in various forms. 
One common issue is the failure to return results that satisfy the structured format required for classification, even after several retry attempts.
Additionally, these models may only partially address the queries posed to them, failing to answer all the target questions comprehensively in the multi-label classification, returning predictions for some labels but not all.
These limitations highlight the ongoing challenges in developing language models that can reliably interpret and respond to complex human inquiries in a consistently accurate and contextually appropriate manner.

We examine the model failure rate in respect of the `format error' failure and `incomplete prediction' failure. 
To reduce the rate of such failures across models, we test the notion of \textit{format demonstrations}, which (unlike few-shot examples) are hypothetical outputs (described further in the methodology).

\begin{figure}[h]
    \centering
    \includegraphics[width=\linewidth]{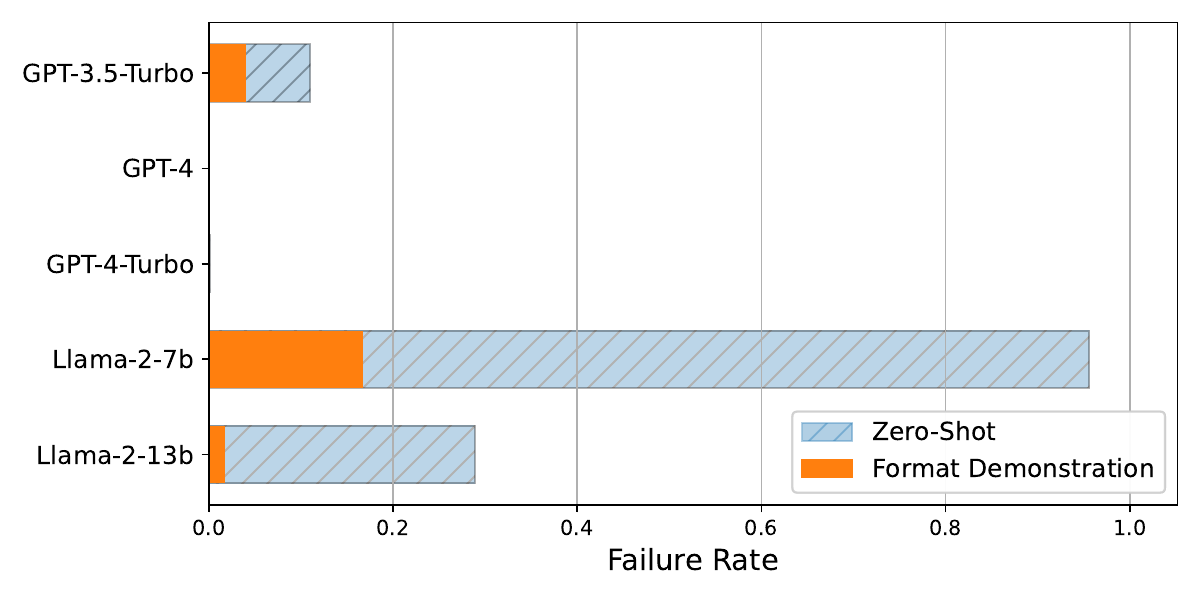}
    \caption{\textbf{Mean failure rates in single-pass classification for various models, under both zero-shot and with format demonstrations.} Format demonstrations vastly reduce the failure rates in all models. GPT-4* models have a near-zero failure rate due to their strength in controlability for respecting instructed output formats. Failures occur due to formatting errors or missing labels, described further in Appendix A.}
    \label{fig:failure_rates}
\end{figure}

\Cref{fig:failure_rates} shows the failure rate of the zero-shot setting, where there's no example provided in the input message for the model to follow, as well.

Format demonstrations substantially reduce the failure rates across all models.
Interestingly, the larger models seem to produce format errors less, potentially due to their demonstrated ability to handle longer contexts.
Llama models incur a significant failure rate, leaving room for further improvements beyond simple format demonstrations.

In order to avoid remaining failures, multiple retries have to be implemented.
When issuing retries, cost and temperature hyperparameter settings are important considerations.
Retrying must be done with temperatures of above 0 to avoid deterministic outputs.
However, temperatures of 0 outperform higher temperatures in all experiments we ran.

\subsection{Comparison of Prompting Strategies}
\begin{figure}[h]
    \centering
    \includegraphics[width=\linewidth]{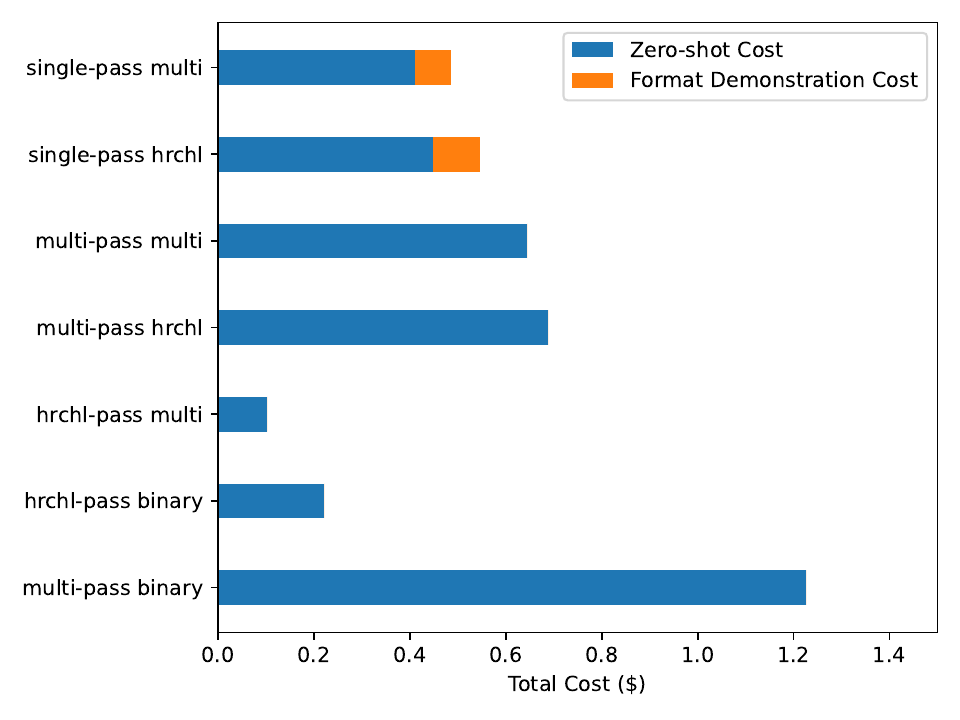}
    \caption{\textbf{Inference costs for each prompting strategy with format demonstration}. Despite using the same model, costs can vary massively (multi-pass binary is 9.4x more expensive than hrchl-pass multi). Costs of single-pass strategies are higher due to needing format demonstrations to reduce the failure rate to a reasonable level, with GPT-3.5-Turbo for example (\Cref{fig:failure_rates}). Overall, performing hierarchical passes in small groups of labels is the cheapest prompting strategy by far, while binary labeling (seeing only one label at a time) is the most expensive. Cost is given for the whole dataset of 200 examples.}
    \label{fig:cost_per_task}
\end{figure}

We compare all prompting strategies described in the methodology.
Prompting strategies use varying amounts of requests and tokens (\Cref{appendix:token}), and therefore vary in cost.
\Cref{fig:cost_per_task} shows the cost of each prompting strategy.
For single-pass options, where all labels are passed to the model in a single request, format demonstrations were needed to reduce failure rates to reasonable levels and to avoid excessive retries.
The added cost of this format demonstration (in orange) is non-negligible; it comes from the added input tokens in each requests and is therefore lesser for single-pass multi than single-pass hrchl, which infers parent labels by the presence of their child-labels.
For all other models, format demonstrations were less cost effective than issuing retries on failures.

Ultimately, performing logic-based hierarchical passes over the label structure is much more efficient for reducing cost than other options.
This is somewhat surprising for hierarchical passes of binary labeling, since each request made to the model only returns a single label out of the whole taxonomy.
Higher frequencies of predicted positives would increase this cost.

\begin{table}[h]
\caption{\textbf{Mean number of passes and tokens required to fully label an example with each prompting strategy}. Multi-pass options split the prediction task into smaller chunks and request predictions from the model in several passes. Input and Output Tokens below describe the total number of tokens to fully label an example passage with the taxonomy of concerns. Results show that multi-pass binary is the least efficient prompting strategy in input tokens. This is because the same prompt instructions introducing the task have to be passed 24 times to the model.}
    \begin{adjustbox}{width=\columnwidth}
    \centering
    \begin{tabular}{lrrr}
    \toprule
    Prompting & Passes & Input Tokens & Output Tokens\\
    \midrule
    single-pass multi  & 1.00 &1066.4 & 255.1\\
    single-pass hrchl &1.00 &1179.4& 319.2\\
    multi-pass multi & 5.00&1851.9& 256.7\\
    multi-pass hrchl & 5.00&1945.6 & 317.5\\
    hrchl-pass multi & 1.29&506.5 & 90.6\\
    hrchl-pass binary & 5.85&1013.1 &6.7 \\
    multi-pass binary &24.00 &4100.4 & 24.8\\
    \bottomrule
    \end{tabular}
    \end{adjustbox}

    \label{appendix:token}

\end{table}

We then examine the performance of these prompting strategies.
Two competing hypotheses are relevant: 
    (1) Allowing the model to focus on fewer target labels at a time reduces noise from other predictions, and increases performance due to better allocation of attention.
    (2) Giving the model multiple target labels at a time has synergistic qualities similar to those explored in multi-task learning (potentially due to the correlated nature of the target labels).

 \begin{table}[h]
    \caption{\textbf{F1 Score (\%) by each model under various prompting strategies}. GPT-4 scores better than GPT-4-Turbo on average with an mean 76.7\% F1 score compared with GPT-4-Turbo's 75.8\% across all prompting strategies. GPT-4 also achieves the best result under a multi-pass binary prompting strategy at 78.65\%. Values are the mean score from two runs with temperature 0 and 1.}
        \centering
        \small
        \begin{tabular}{lccc}
        \toprule
        Prompting & GPT-3.5-Turbo & GPT-4 & GPT-4-Turbo \\
        \midrule
        single-pass multi &65.63 &76.41 &77.03 \\
        single-pass hrchl &63.95 &77.05 & \textbf{78.33}\\
        multi-pass multi & \textbf{67.51}&76.99& 75.83\\
        multi-pass hrchl & 64.25& 75.02& 72.87\\
        hrchl-pass multi & 59.73&78.56 & 74.70\\
        hrchl-pass binary & 57.59& 74.19& 74.22\\
        multi-pass binary &61.81&\textbf{78.65} &77.45 \\
        \bottomrule
    \end{tabular}
    \label{tab:f1_perf_strategies}
\end{table}

The performances are compared in terms of F1 Score versus the determined gold labels described in \citet{stureborg2023interface}.
We focus on F1 score since different applications downstream from the classification of vaccine concerns in text may care more or less about precision and recall, respectively.
For example, applications that provide interventions may focus on higher precision (to be sure they are intervening on an truly expressed concern) whereas applications interested in measuring the presence of concerns within a population may focus on recall to avoid potentially missing a concern.

\Cref{tab:f1_perf_strategies} describes the performance achieved using each prompting strategy, separated by model.
Multi-pass binary---performing parallel passes of binary target labels---is the most expensive prompting strategy and achieves the best performance in GPT-4. 
However, GPT-4-Turbo achieves a close performance of 78.33\% F1 under the single-pass hrchl scheme, which is much cheaper.

\subsection{Cost-Performance Tradeoff}
\begin{figure}[h]
    \centering
    \includegraphics[width=0.95\linewidth]{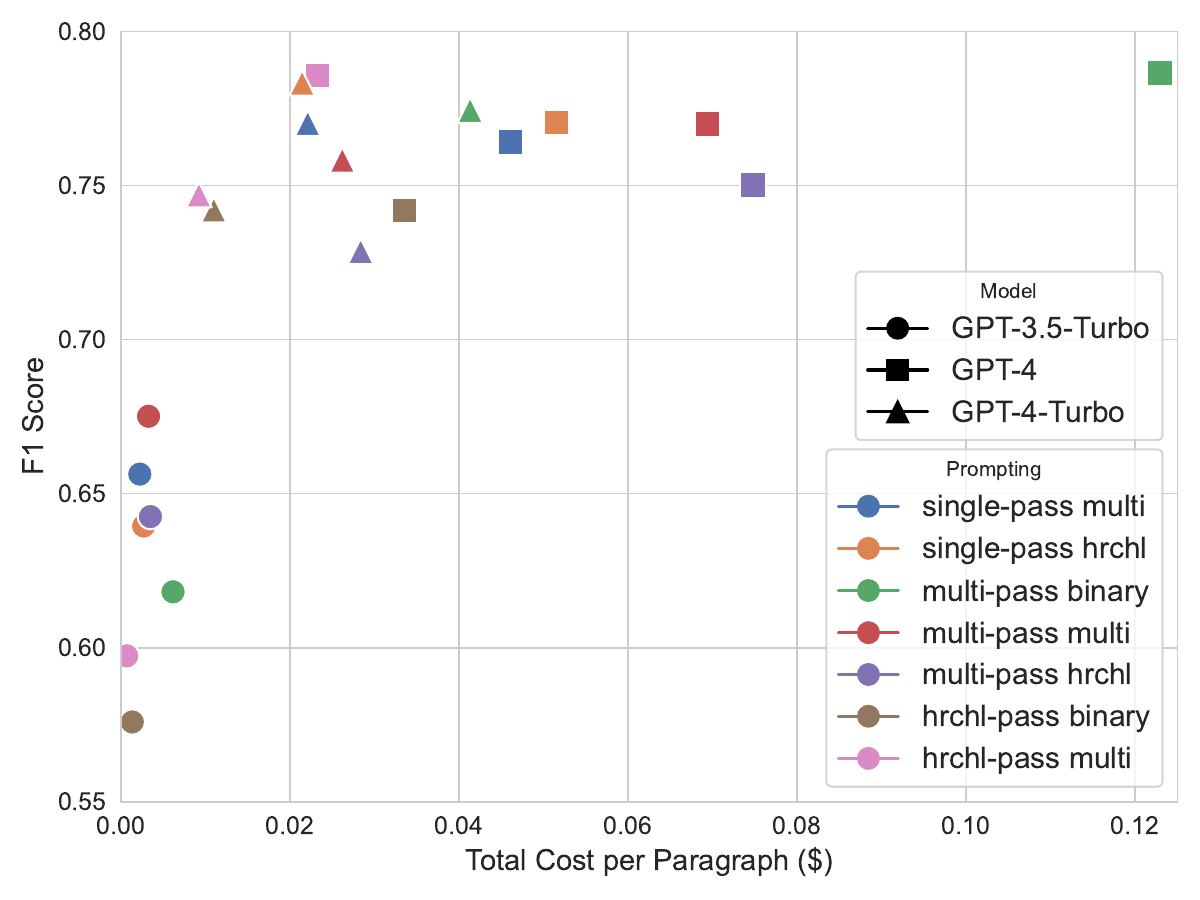}
    \caption{\textbf{Total cost versus performance by model and prompting strategy}. Throughout our experiments, we show a positive relationship between cost of prediction and performance. However, this relationship is largely driven by model cost differences. Yet, the relationship between the cost of prompting strategies and their performance is positive. This could potentially hint that models perform better when focusing on fewer labels per generation.}
    \label{fig:cost_vs_perf}
\end{figure}

Due to the massive scale of text on social media, establishing any form of monitoring system or large-scale data analysis requires great consideration to cost. 
Performance has to be balanced with the cost of the classification system and \Cref{fig:cost_vs_perf}
shows this tradeoff.

Note that while there is a general trend of higher cost classification systems having higher performance, this trend is mostly due to model selection.
Prompting strategies' costs are less correlated with performance.
Therefore, there is an opportunity to select cheaper prompting strategies without sacrificing as much performance.

\Cref{fig:cost_vs_perf} shows several (7) Pareto optimal classification systems. Two of the most obvious are GPT-4-Turbo using single-pass hrchl and GPT-4 using hrchl-pass multi.
These systems are Pareto optimal, since optimizing either cost or performance further must come at the expense of a loss in the other.
If one wants to minimize cost, hrchl-pass multi using GPT-3.5-Turbo is the optimal system, however, this has a performance slightly below 0.60 F1 score.
Conversely, the best-performing system is also the most expensive.

\subsection{Comparison to Crowdsource Worker Performance}
The top performance across our experiments was using GPT-4 with multi-pass binary prompting.
The concern-wise performance is summarized in \Cref{tab:perf_by_concern}.

\begin{table}[h]
    \caption{\textbf{Concern-wise F1 Score (\%) of Human Annotation versus GPT-4}. We compare the best performance of majority vote human annotation using the single-pass hrchl strategy with GPT-4 using multi-pass binary. The human annotation is determined by the majority vote of three crowdsource workers who were presented the passage in a single pass with the full taxonomy for annotation. This is the best prompting strategy as determined by \citet{stureborg2023interface}, although they did not explore binary strategies due to costs. Missing values arise due to either no positive examples in the ground-truth labels (thus a divide by zero in recall) or no positive predictions.}
    \begin{adjustbox}{width=\linewidth}
    \centering
    \small
    \begin{tabular}{lrrrr}
         \toprule
        \textbf{Concern}  & \textbf{GPT-4} & \textbf{Crowdsource Workers} \\
        \midrule
        Issues with Research &  79.57 & 58.06\\
        \hspace{5mm}Lacking Quality &  82.68 & 36.36\\
        \hspace{5mm}Poor Quality &  80.76 & 63.16\\
        \hspace{5mm}Fallible Science &  - & -\\
        \midrule
        Lack of Benefits &  82.13 & 80.00\\
        \hspace{5mm}Imperfect Protection &  87.44 & 70.00\\
        \hspace{5mm}Herd Immunity &  - & -\\
        \hspace{5mm}Natural Immunity &  83.21 & 100.00\\
        \hspace{5mm}Insufficient Risk & 100.00 &  100.00\\
        \hspace{5mm}Existing Alternatives &  89.87 &100.00 \\
        \midrule
        Health Risks &  87.94 & 85.06\\
        \hspace{5mm}Direct Transmission & - & -\\
        \hspace{5mm}Harmful Ingredients & 95.19 & 95.24\\
        \hspace{5mm}Specific Side Effects &  66.86 & 76.67\\
        \hspace{5mm}Dangerous Delivery &  49.24 & 66.67\\
        \hspace{5mm}High-Risk Individuals & 49.37 & 28.57\\
        \midrule
        Disregard of Individual Rights &  79.24 & 70.59\\
        \hspace{5mm}Religious and Ethical Beliefs &  89.87 & 22.22\\
        \hspace{5mm}Right to Autonomy & 89.74 & -\\
        \midrule
        Untrustworthy Actors &  75.76 & 78.26\\
        \hspace{5mm}Incompetence &  55.59 & 100.00\\
        \hspace{5mm}Profit Motives & 89.19 & 80.00\\
        \hspace{5mm}Censorship &  73.96 & 60.00\\
        \hspace{5mm}Conspiracy & 66.21 & 36.36\\
        \midrule
        \textbf{Mean} &\textbf{ 78.75} &\textbf{70.36} \\
        \bottomrule
    
    \end{tabular}
    \end{adjustbox}
    \label{tab:perf_by_concern}
\end{table}

Surprisingly, this performance is substantially better than human-based crowdsource annotation baselines.
In crowdsourcing, the best-achieved performance is 70.4\% F1 score using a single-pass hrchl annotation interface and combining three worker's annotations using majority vote.
This strategy is much more expensive than LLM-based annotation, and it outperformed by +0.08 F1 score at a total 78.7\% F1.

Breaking the task down further, this seems to be due to LLMs high consistency on easy examples.
\Cref{tab:easy_vs_not} shows the performance of GPT-4 on the VaxConcerns dataset broken down by example difficulty.
The categories of difficulty are explained in the Methodology.

It is possible that the high performance of LLM labeling is due to their consistency on easy examples, whereas humans can lose focus or make careless mistakes during annotation such as instantly assuming a label applies due to keywords being present.
This is supported by the results of \cite{stureborg2023interface}, which found that crowdsource workers showed no measurable improvement on the easier examples.
Further analysis and experimentation are required to determine the cause of this difference.

\begin{table}[h]
    \caption{\textbf{F1 Score (\%) by difficulty, GPT-4 at temperature 0}. Easy examples are those where experts from \citet{stureborg2023interface} had unanimous agreement in the first round of labeling. Note that the performance on this subset of examples is particularly high, 84.3\% for single-pass hrchl prompting. This indicates a subset of the dataset is easier for GPT-4, whereas the non-easy partition has good performance but leaves room for substantial improvements.}
    \centering
    \small
    \begin{tabular}{lrr|r}
    \toprule
    Prompting & Easy & Non-Easy & Overall\\
    \midrule
    single-pass multi  & 80.84 & 64.68 & 78.31 \\
    single-pass hrchl  & \textbf{84.27} & 65.13 & 77.62 \\
    multi-pass multi   & 75.76 & 64.59 & 77.15 \\
    multi-pass hrchl   & 75.76 & 66.81 & 75.72 \\
    hrchl-pass multi   & 82.38 & \textbf{66.97} & 78.63 \\
    hrchl-pass binary  & 81.41 & 58.57 & 73.80 \\
    multi-pass binary  & 81.62 & 63.60 & \textbf{78.75} \\

    \bottomrule
    \end{tabular}
    \label{tab:easy_vs_not}
\end{table}
\section{Limitations and Future Work}

\textit{Limited Model Selection}.
We make use of both instruction tuned chatGPT models, as well as two Llama-2 variants.
Due to constraints in the project scope, time, and performance of Llama-2 (especially on format errors), we performed most of the experiments using the GPT models exclusively.
This is problematic for a few reasons.
These models are opaque systems, with little insight into the training procedures and training data, pre- and post-processing that is carried out, and other manipulations by OpenAI.
However, this is a difficult limitation to entirely get rid of due to the strength of this class of models.
GPT models are SOTA across many tasks in NLP, and ignoring them would be to ignore the currently most relevant models in the field.

\textit{Dataset Size and Focus}.
We use the dataset provided by \citet{stureborg2023interface}.
This dataset only contains 200 passages and 4,800 passage-label pairs for evaluation.
This reduces the statistical power of any findings our results show.
Our experiments are therefore preliminary and we believe helpful, but further experiments have to be carried out to verify the trends we have uncovered.
Further, the dataset focuses exclusively on anti-vaccination blog text.
This is very different from the domain of social media text both in style and content, and we can therefore not know whether or the performance will generalize out of this domain.

\textit{Limited to a Single Taxonomy}.
In our experiments, we only investigate a single taxonomy as the target for hierarchical multi-label classification.
It is possible that results will vary if introducing taxonomies of different sizes or even domains.
That being stated, the core motivation for this work is to allow for the detection of vaccine concerns in online text.
For this purpose, VaxConcerns is the best option due to its crowdsource viability, high quality evaluation set, and the fact that it is disease-agnostic and therefore robust to new viruses/diseases that may arise in a future global health crisis.
We therefore believe furthering this resource is of value to the healthcare community.

\section{Conclusion}

In this work, we have addressed the task of detecting vaccine concerns in online discourse using large language models (LLMs) in a zero-shot setting.
We explore cost-performance trade-offs of different prompting strategies and benchmark LLMs on hierarchical multi-label classification of the VaxConcerns taxonomy, showing that it is possible to trade performance for cost.
We map out this tradeoff curve and find that GPT-4 with multi-pass binary scores best with an F1 score of 78.65\%.
This is also higher than the best crowdsource worker annotation scheme which achieves 70.36\% F1.
We show the effectiveness of introducing format demonstrations to avoid output format failures, vastly reducing failure rates across all models.

\section{Acknowledgments}
We thank Jun Yang for his advise on research directions, and the anonymous reviewers for their feedback. This work was supported by NSF award IIS-2211526 and an award from Google.


\setcounter{secnumdepth}{3} 
\appendix

\bibliographystyle{aaai}
\bibliography{bibliography}
\end{document}